\documentclass[10pt,twocolumn,letterpaper]{article}

\usepackage{cvpr}
\usepackage{times}
\usepackage{epsfig}
\usepackage{graphicx}
\usepackage{amsmath}
\usepackage{amssymb}

\usepackage{mathrsfs}
\usepackage{ifpdf}
\usepackage{cite}
\usepackage{multirow}
\usepackage{amsfonts}
\usepackage{amsthm}
\usepackage{algorithm}
\usepackage{algorithmic}
\usepackage{array}
\usepackage[marginal]{footmisc}
\usepackage[caption=false,font=footnotesize]{subfig}
\usepackage{stfloats}
\usepackage[usenames]{color}
\usepackage{multirow}
\usepackage{url}
\usepackage{booktabs}
\usepackage{enumerate}
\usepackage{gensymb}
\usepackage{bbm}
\usepackage{stmaryrd}
\usepackage{float}
\usepackage{textcomp}
\usepackage{longtable}
\usepackage{boxedminipage}
\usepackage{chemarrow}
\usepackage{extarrows}
\usepackage{array}
\usepackage{bm}
\makeatletter
\newcommand{\thickhline}{%
    \noalign {\ifnum 0=`}\fi \hrule height 1pt
    \futurelet \reserved@a \@xhline
}
\newcolumntype{"}{@{\hskip\tabcolsep\vrule width 1pt\hskip\tabcolsep}}
\makeatother

\DeclareMathOperator*{\argmin}{arg\,min}
\usepackage[pagebackref=true,breaklinks=true,letterpaper=true,colorlinks,bookmarks=false]{hyperref}

\cvprfinalcopy % *** Uncomment this line for the final submission

\def\cvprPaperID{397} % *** Enter the CVPR Paper ID here

\ifcvprfinal\fi
\begin{document}

%%%%%%%%% TITLE
\title{Zigzag Learning for Weakly Supervised Object Detection}
\author{Xiaopeng Zhang$^{1}$ \, Jiashi Feng$^1$ \, Hongkai Xiong$^2$ \, Qi Tian$^3$\\\\
$^1$ \normalsize {National University of Singapore}\,
$^2$ \normalsize {Shanghai Jiao Tong University} \,
$^3$ \normalsize {University of Texas at San Antonio}\\
{\tt\small \{elezxi,elefjia\}@nus.edu.sg} \,
{\tt\small xionghongkai@sjtu.edu.cn} \, {\tt\small qitian@cs.utsa.edu}
}

\maketitle
\thispagestyle{empty}

%%%%%%%%% ABSTRACT
\begin{abstract}
  %Current state-of-the-art detection models require object-level annotations at training stages, but collecting them is tedious and time-consuming.
  This paper addresses weakly supervised object detection with only image-level supervision at training stage. Previous approaches train detection models with entire images all at once, making the models prone to being trapped in sub-optimums due to the introduced false positive examples. Unlike them, we propose a zigzag learning strategy to simultaneously discover reliable object instances and prevent the model from overfitting initial seeds. Towards this goal, we first develop a criterion named mean Energy Accumulation Scores (mEAS) to automatically measure and rank localization difficulty of an image containing the target object, and accordingly learn the detector progressively by feeding examples with increasing difficulty. In this way, the model can be well prepared by training on easy examples for learning from more difficult ones and thus gain a stronger detection ability more efficiently. Furthermore, we introduce a novel masking regularization strategy over the high level convolutional feature maps to avoid overfitting initial samples. These two modules formulate a zigzag learning process, where progressive learning endeavors to discover reliable object instances, and masking regularization increases the difficulty of finding object instances properly. We achieve $47.6\%$ mAP on PASCAL VOC 2007, surpassing the state-of-the-arts by a large margin.
\end{abstract}

%%%%%%%%% BODY TEXT
%the detection performance under fully supervised paradigm have reached over $88\%$. However,
%There is a long way to go comparing with fully supervised detections.

%based on the observation that recognition usually focused on object parts instead of the whole object. We propose to sample positive instances according to the saliency-based scores.
%The advantages of using energy accumulation
\section{Introduction}
%As a fundamental problem in computer vision, object detection has experienced renewed development due to the success of convolutional neural networks \cite{krizhevsky2012imagenet}.
Current state-of-the-art object detection performance has been achieved with a fully supervised paradigm. However, it requires a large quantity of high-quality object-level annotations (\emph{i.e.}, object bounding boxes) at training stages \cite{girshick2015fast}, \cite{liu2016ssd}, \cite{Redmon_2016_CVPR}, which are very costly to collect. %, especially for large scale datasets.
Fortunately, the prevalence of image tags allows search engines to quickly provide a set of images related to the target category \cite{niu2015visual}, \cite{vijayanarasimhan2008keywords}, making image-level annotations much easier to acquire. Hence it is more appealing to learn detection models from such weakly labeled images. %, which would lift the generalizability of trained models to new object classes without human generated object-level annotations.
In this paper, we focus on object detection under a weakly supervised paradigm, where \emph{only image-level labels} indicating the presence of an object are available during training.

\begin{figure}[t]
  \centering
  % Requires \usepackage{graphicx}
  \includegraphics[width=0.46\textwidth]{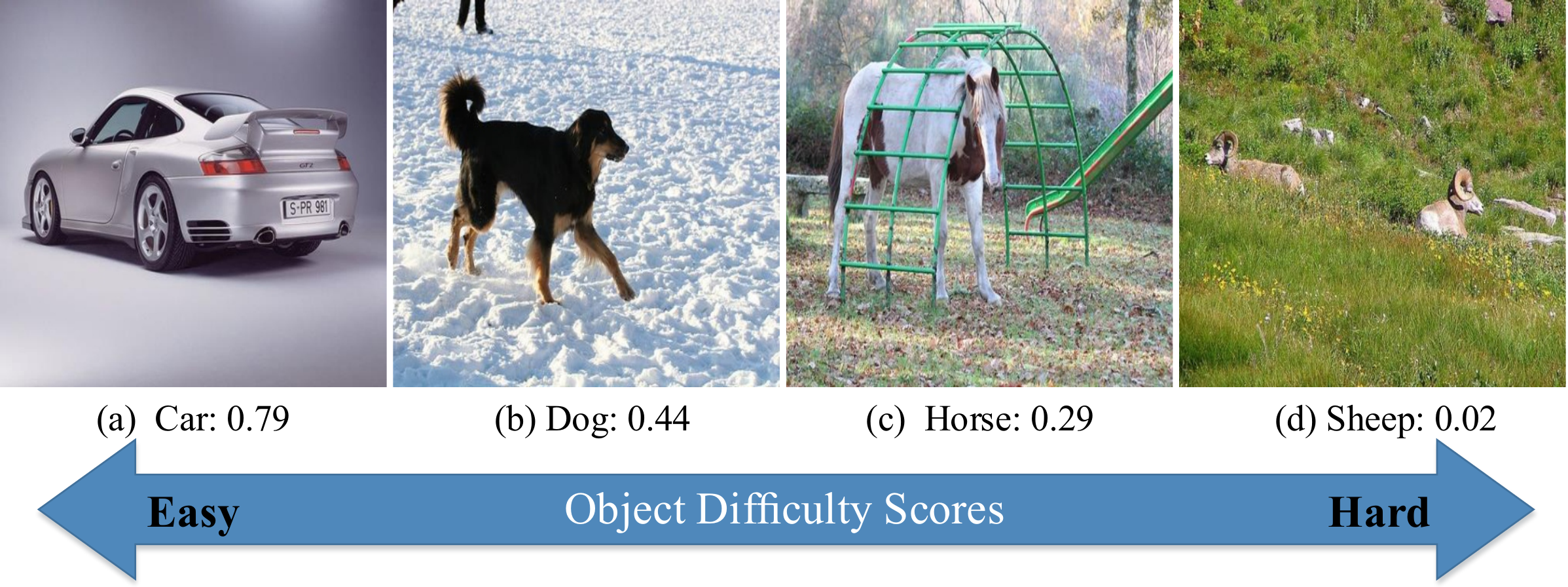} \\
  \caption{Object difficulty scores predicted by our proposed mEAS. Higher scores indicate the object is easier to localize. This paper proposes a zigzag learning based detector to progressively learn from object instances in the order according to mEAS, with a novel masking regularization to avoid overfitting initial samples.} \label{motivation}
  \vspace{-0.3cm}
\end{figure}

The main challenge in weakly supervised object detection is how to disentangle object instances from the complex backgrounds. Most previous methods model the missing object locations as latent variables, and optimize them via different heuristic methods \cite{li2016image}, \cite{song2014weakly}, \cite{wang2014weakly}. Among them, a typical solution is alternating between model re-training and object re-localization, which shares a similar spirit with Multiple Instance Learning (MIL)~\cite{cinbis2014multi}, \cite{Li_2016_CVPR}, \cite{Tang_2017_CVPR}. Nevertheless, such optimization is non-convex and easy to get stuck in local minimums if the latent variables are not properly initialized. Then mining object instances with only image-level labels becomes a classical chicken-and-egg problem: without an accurate detection model, object instances cannot be discovered, while an accurate detection model cannot be learned without appropriate object examples.
%As a result, current weakly supervised detection approaches still suffer limited performance, which are far below that of the fully supervised counterparts.
%In this paper, we raise two issues, \emph{1) how to obtain reliable positive instances with only image-level labels? 2) how to avoid overfitting for model training?}
%However, images are suffering different degrees of illumination and view variations, and sometimes are partially occluded or within cluttered scenes. As a results, some objects are inherently difficult to localize, it is suboptimal to apply the same object mining algorithm for all the images.
% To address this issue, a typical solution is based on an iterative, multiple instance learning (MIL) strategy \cite{cinbis2014multi}, \cite{Li_2016_CVPR}, \cite{Tang_2017_CVPR}, in which an image is treated as a bag of instances containing potential object instances. The instance mining process is performed by iteratively alternating between training detection models and re-localizing new object instances. However, MIL is sensitive to initialization and easy to get stuck in local minima. As a result, current weakly supervised detection methods suffer limited performance, which are far below that of the fully supervised counterparts.
%An ideal situation is that these labeling methods would produce detection performance approaching to the fully supervised one, while it is not the case.

%considers the order of training images to be processed in learning a detection model.

To solve this problem, this paper proposes a \emph{zigzag} learning strategy for weakly supervised object detection, which aims at mining reliable object instances for model training, and meanwhile avoiding getting trapped in local minimums. As our first contribution, different from previous works which perform model training and object re-localization over the entire images all at once \cite{Li_2016_CVPR}, \cite{Tang_2017_CVPR}, \cite{jie2017deep}, we progressively feed the images into the learning model in an easy-to-difficult order \cite{kumar2010self}. To this end, we propose an effective criterion named mean Energy Accumulated Scores (mEAS) to automatically measure the difficulty of an image containing the target object, and progressively add samples during model training. % The motivation is that objects are within different background clutters, which makes some objects intrinsically hard to localize.
As shown in Fig. \ref{motivation}, \textit{car} and \textit{dog} are simpler to localize while \textit{horse} and \textit{sheep} are more difficult. Intuitively, ignoring this discrepancy of object difficulty in localization would inevitably include many poorly localized samples, which deteriorates the trained model. On the other hand, processing easier images in the initial stages leads to better detection models, which in turn increases the probability of successfully localizing objects in difficult images.

% Instead of training detection models with the entire images all at once, we propose a self paced learning strategy. We demonstrate that the order in which images are processed is important in weakly supervised detection.
% As a results, the proposed method formulates a self paced learning strategy \cite{kumar2010self}, in which
% \emph{i.e.,} finding reliable positive instances for model training, and avoiding overfitting in re-training.
%  Another critical issue in weakly supervised detection is that models are prone to focusing on the most discriminative parts, instead of the whole object, producing conservative object bounding boxes.
Due to lack of object annotations, the mined object instances inevitably include false positive samples. Current approaches \cite{Li_2016_CVPR}, \cite{Tang_2017_CVPR} simply treat these pseudo annotations as ground truth, which is suboptimal and easy to overfit the initial seeds. This is especially true for a deep network due to its high fitting capacity. As our second contribution, we design a novel masking strategy over the last convolutional feature maps, which randomly erases the discriminative regions during training. It prevents the model from concentrating on part details at earlier training, and induces the network to focus more on those less discriminative parts at current  training. In this way, the model is able to discover more integrated objects as desired. Another advantage is that the proposed masking operation introduces many random occluded samples, which can be treated as data augmentation and enhances the generalization ability of the model.

Integrating the progressive learning and masking regularization formulates a zigzag learning process. The progressive learning endeavours to discover reliable object instances in an easy-to-difficult order, while the masking strategy increases the difficulty in a way favorable of object mining via introducing many random occluded samples. These two adversarial modules boost each other, and benefit both object instance mining and reducing model overfitting risks. The effectiveness of zigzag learning has been validated experimentally. On benchmark dataset PASCAL VOC 2007, we achieve an accuracy of $47.6\%$ under weakly supervised paradigm, which surpasses the-state-of-the-arts by a large margin.
To sum up, we make following contributions.

$\bullet$ We propose a new and effective criterion named mean Energy Accumulated Scores (mEAS) to automatically measure the difficulty of an image w.r.t. localizing a specific object. Based on mEAS, we train detection models via an easy-to-hard strategy. This kind of progressive learning is beneficial to finding reliable object instances especially for the difficult images.

$\bullet$ We introduce a feature masking strategy during an \emph{end-to-end} model learning, which not only forces the network to focus on less discriminative details during training, but also avoids model overfitting via introducing random occluded positive instances. Integrating these two components gives a novel zigzag learning method and achieves state-of-the-art performance for weakly supervised object detection.

\begin{figure*}[t]
  \centering
  % Requires \usepackage{graphicx}
  \includegraphics[width=1.0\textwidth]{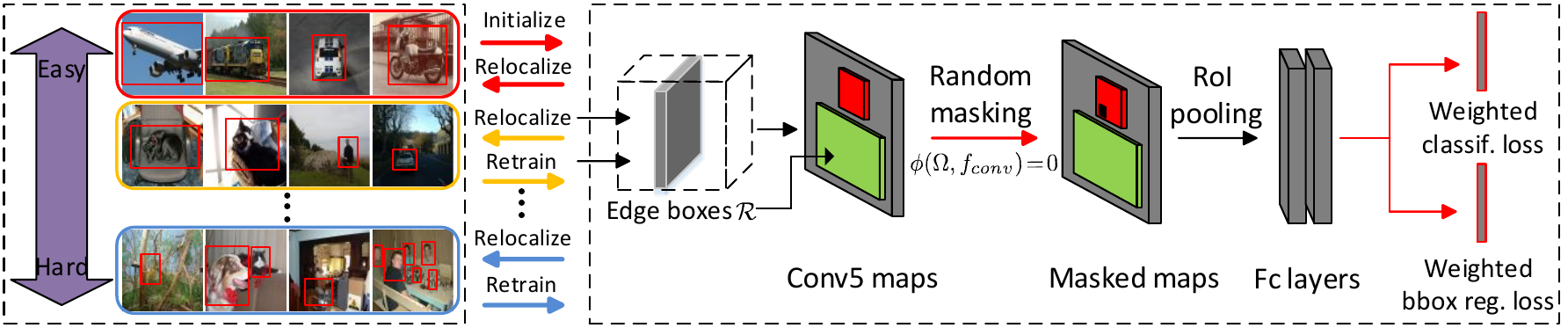} \\
   \vspace{0.1cm}
  \caption{Architecture of our proposed zigzag detection network. We first estimate the image difficulty with mean Accumulated Energy Scores (mEAS), organizing training images in an easy-to-difficult order. Then we introduce a masking strategy over the last convolutional feature maps of fast RCNN framework, which enhances the generalization ability of the model.} \label{flow_chart}
  \vspace{-0.1cm}
\end{figure*}

\section{Related Works}
%Most of recent works in weakly supervised detection follow a similar framework. At first, a set of region proposals are extracted, then the object regions are selected using different region mining methods.
%Previous works related to weakly supervised detection mainly focus on two tasks: latent object-level instance mining, and detection model optimization. In the following, we organize our discussion within the above aspects.
%Object detection has experienced renewed success due to the introduction of CNN. The pioneer work of Girshick \emph{et. al} \cite{girshick2014rich}. Recently, the proposed SSD \cite{liu2015ssd} and YOLO \cite{Redmon_2016_CVPR} frameworks enable us to obtain comparative performance at the cost of tens of frames in one second, and enable us moving to the area of real-time object detection.
Our method is related with two fields: 1) image difficulty evaluation; 2) weakly supervised detection. %In the following, we organize our discussion with the above aspects.

\textbf{Evaluating image difficulty.} Little literature has been devoted to evaluating the difficulty of an image. A preliminary work in \cite{tudor2016hard} estimates the image difficulty via analyzing some low-level cues such as edges, segments, and objectness scores. Similarly,  \cite{shi2016weakly} assumes that image difficulty is most related with the object size, and builds a regression model to estimate the object size in an image. However, it needs extra object size annotations for training the regressor. In contrast, we propose an easy-to-compute criterion named mean Accumulated Energy Scores (mEAS) to automatically measure the difficulty of an image. The advantage is that the criterion is based on the network itself, and free of human interpretation. %The experimental results demonstrate how our proposed network oriented metric boosts the detection model.

\textbf{Weakly supervised detection.} It is intuitive to mine object instances from weakly labeled images \cite{song2014weakly}, \cite{wang2014weakly}, \cite{Li_2016_CVPR}, and follow the pipeline of fully supervised detection based on the mined objects. Our proposed method is most related with \cite{cinbis2014multi}, \cite{Li_2016_CVPR}, \cite{Tang_2017_CVPR}, which try to obtain reliable object instances via an iterative updating strategy. However, these methods either detach the feature extraction and model training into separate steps \cite{cinbis2014multi}, \cite{Li_2016_CVPR}, or simply utilize the high representation ability of CNN without considering model overfitting \cite{Tang_2017_CVPR}, which results in limited performance. Comparatively, we integrate model training and object mining into a unified framework, and propose a zigzag learning strategy to improve the generalization ability of the model. These modifications enable us to achieve superior detection accuracy under the weakly supervised paradigm.

Our method is also related with  \cite{oquab2015object}, \cite{Bilen_2016_CVPR}. Oquab \emph{et al.} \cite{oquab2015object} proposed a weakly supervised object localization method by explicitly searching over candidate object locations at different scales during training. However, their localization result is limited since it only returns a center point for an object, not the tight bounding box. Bilen \cite{Bilen_2016_CVPR} \emph{et al.} proposed to model image-level loss as the accumulated scores over regions and performed detection based on the region scores. Nevertheless, this network is modeled as classification loss, which makes the detection model easily focus on object parts rather than the whole objects.
\section{Method}
In this section, we elaborate on the proposed zigzag learning based weakly supervised detection model. Its overall architecture consists of three modules, as shown in Fig.~\ref{flow_chart}. The first module estimates image difficulty automatically via a backbone network~\cite{bency2016weakly} trained with only image-level labels. The second module progressively adds samples to network training in an ascending order based on image difficulty. Third, we incorporate convolutional feature masking into model training to regularize the high responsive patches during previous training and enhance the generalization ability of the model. In the following, we discuss these modules in details.

\subsection{Estimating Image Difficulty}
%Mining object instances with only image-level labels is a classical chicken-and-egg problem: without an accurate detection model, object instances cannot be discovered, while an accurate detection model cannot be learned without appropriate object exemplars.
Images differ in their difficulty for localization, which comes from factors such as object size, background clutter, number of objects, and partial occlusion. For subjective evaluation, image difficulty can be quantified as the time needed by a human to determine the actual position of a given class \cite{tudor2016hard}. However, this brings about extra human efforts. In this subsection, we evaluate the image difficulty via diagnosing its localization outputs.

\textbf{WSDDN framework.} Our method needs a pretrained model to diagnose the localization outputs of an image. Without loss of generality, we use WSDDN \cite{Bilen_2016_CVPR} as the baseline network, for its effectiveness and implementation convenience. WSDDN explicitly models image-level classification loss via aggregating region proposal scores. Specifically, given an image $x$ with region proposals $\mathcal{R}$, and image level labels $y\in\{1,-1\}^C$, where $y_c\!=\!1$ ($y_c\!=\!-1$) indicates the presence (absence) of an object class $c$. Denote the outputs of $fc_{8C}$ and $fc_{8R}$ layer as $\phi(x, fc_{8C})$ and $\phi(x, fc_{8R})$, respectively, which are with size $C \times |\mathcal{R}|$. Here, $C$ represents the number of categories and $|\mathcal{R}|$ denotes the number of regions. The score of region $r$ corresponding to class $c$ is the dot product of the two fully connected layers $\phi(x, fc_{8C})$ and $\phi(x, fc_{8R})$, normalized at different dimensions:
\begin{equation}\label{region_score}
 x_{cr} = \frac{e^{\phi^{cr}(x,fc_{8C})}}{\sum_{i=1}^{C}e^{\phi^{ir}(x,fc_{8C})}} .*\frac{e^{\phi^{cr}(x,fc_{8R})}}{\sum_{j=1}^{|\mathcal{R}|}e^{\phi^{cj}(x,fc_{8R})}}.
\end{equation}
Based on the region-level score $x_{cr}$, the probability output $y$ w.r.t. category $c$ at image-level is defined as the sum of a series of region-level scores:
\begin{equation}\label{image-prob}
 \phi^c(x,w_{cls}) = \sum_{j=1}^{|\mathcal{R}|}x_{cj},
\end{equation}
where $w_{cls}$ denotes the non-linear mapping from input $x$ to classification stream output. This network is back-propagated via a binary log image-level loss, denoted as
\begin{equation}\label{binary-log}
L_{cls}(x,y)= \sum_{i=1}^C\log(y_i(\phi^i(x,w_{cls})-1/2)+1/2),
\end{equation}
and is able to automatically localize the regions which contribute most to the image level scores.

\begin{figure*}[t]
  \centering
  % Requires \usepackage{graphicx}
  \includegraphics[width=1.0\textwidth, height=4.5cm] {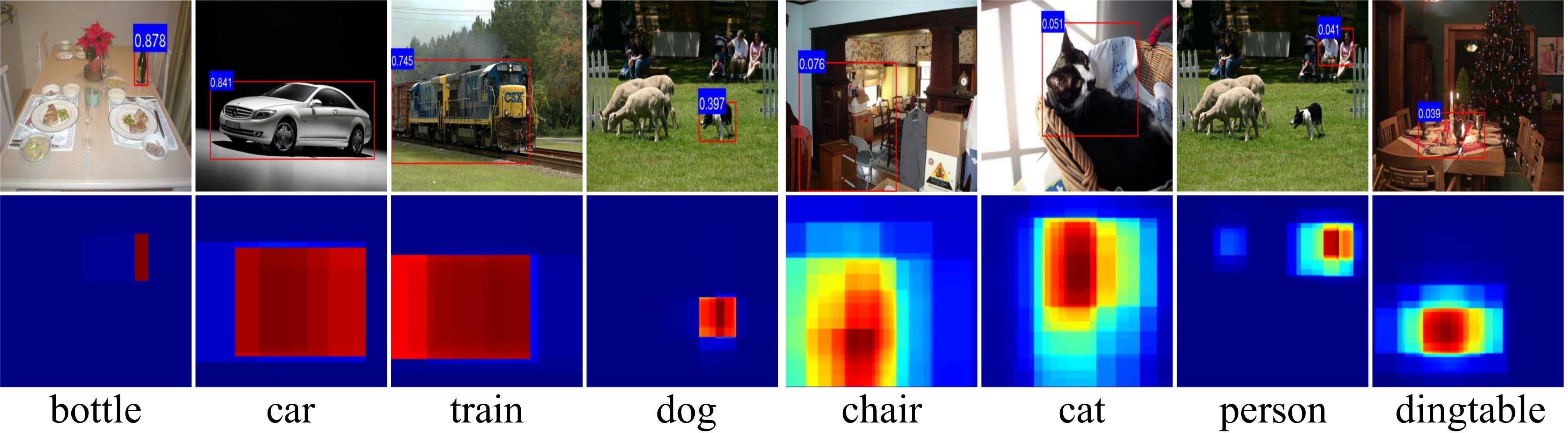} \\
   %\vspace{0.1cm}
  \caption{Example image difficulty scores by the proposed mEAS metric. Top row: mined object instances and mEAS. Bottom row: corresponding object heat maps produced by Eq. (\ref{heatmap}). Best viewed in color.} \label{image_heat}
  \vspace{-0.1cm}
\end{figure*}

\textbf{Mean Energy Accumulated Scores (mEAS).} Benefiting from the competitive mechanism, WSDDN is able to pick out the most discriminative details for classification. These details sometimes fortunately correspond to the whole object, but in most cases only focus on object parts. We observe that the successfully localized objects usually appear in relatively simple, uniform background with only a few objects in the image. In order to pick out images that WSDDN localizes successfully, we propose an effective criterion named mean Energy Accumulated Scores (mEAS) to quantify the localization difficulty of each image.

If the target object is easy to localize, the regions that contribute most to the classification scores should be highly concentrated.
To be specific, given an image $x$ with labels $y\in\{1,-1\}^C$, for each class $y_c\!=\!1$, we sort the region scores $x_{cr} \, (r \in \{1,..., |\mathcal{R}|\})$ in a descending order, and obtain the sorted list $x_{cr'}$, where $r'$ is a permutation of $\{1,..., |\mathcal{R}|\}$. Then we compute the accumulated scores of $x_{cr'}$ to obtain a monotonically increasing list $X_c \in {R^{|\mathcal{R}|}}$, with each dimension denoted as
\begin{equation}\label{acc_scores}
  X_{cr} = \sum_{j=r'(1)}^{r'(j)} x_{cj}/\sum_{j=1}^{|\mathcal{R}|} x_{cj}.
\end{equation}
$X_c$ is in the range of $[0 \,\, 1]$ and can be regarded as an indicator depicting the convergence degree of the region scores. If the top scores only focus on a few regions, then $X_c$ converges quickly to 1. In this case, WSDDN is easy to pick out the target object.

Inspired by the precision/recall metric, we introduce Energy Accumulated Scores (EAS) to quantify the convergence of $X_c$. EAS is inversely proportional to the minimal number of regions needed to make $X_c$ above a threshold $t$,
\begin{equation}\label{eas}
  EAS(X_c, t)= \frac{X_{cj_{[t]}}}{j_{[t]}},\ \ \ \ j_{[t]}= \argmin_j X_{cj} \geq t.
\end{equation}
It is obvious that a larger $EAS(X_c, t)$ means that fewer regions will be needed to reach the target energy. Finally, we define the mean Energy Accumulated Scores (mEAS) as the mean scores at a set of eleven equally spaced energy levels $[0,0.1,...,1]$:
\begin{equation}\label{meas}
  mEAS(X_c)= \frac{1}{11} \sum_{t \in \{0,0.1,...,1\}} EAS(X_c, t).
\end{equation}
%note that in our definition, the image difficulty scores are category specific, \emph{i.e.,} we compute mEAS of an image for different classes, respectively.

\textbf{Mining object instances.} Once we obtain the image difficulty, the remaining task is to mine object instances from the images. A natural way is to directly choose the top scored region as the target object, which is used for localization evaluation in \cite{bency2016weakly}. However, since the whole network is trained with classification loss, which makes high scored regions tend to focus on object parts rather than the whole objects. To relieve this issue, we do not optimistically consider the top scored region to be accurate enough. In contrast, we consider them to be accurate enough as soft voters. To be specific, we compute the object heat map $H^c$ for class $c$, which collectively returns the confidence that pixel $p$ lies in an object, \emph{i.e.,}
\begin{equation}\label{heatmap}
  H^c(p)=\sum_rx_{cr}D_r(p)/Z,
\end{equation}
where $D_r(p)\!=\!1$ when the $r$-th region proposal contains pixel $p$, and $Z$ is a normalization constant such that $\max H^c(p)\!=\!1$. We binarize the heat map $H^c$ with threshold $T$ (set as $0.5$ in all experiments), and choose the tightest bounding box that encloses the largest connect component as the mined object instance.

\textbf{Analysis of mEAS.} mEAS is an effective criterion to quantify the localization difficulty of an image. Fig. \ref{image_heat} shows some image difficulty scores from mEAS on PASCAL VOC 2007 dataset, together with the mined object instances (top row) and object heat maps (bottom row). It can be seen that images with higher mEAS are easy to localize, and the corresponding heat maps exhibit excellent spatially convergence characteristics. In contrast, images with lower mEAS are usually hard to localize, and the corresponding heat maps are divergent. Comparing with the region scores in Eq.~(\ref{region_score}), mEAS is especially effective in filtering out the inaccurate localizations in these two cases:

$\bullet$ The top scored regions only focus on part of the object. This usually occurs on non-rigid objects such as \textit{cat} and \textit{person} (see the 6th column in Fig. \ref{image_heat}). In this case, the less discriminative parts make the heat maps relatively divergent, and thus lower the mEAS.

$\bullet$ There exist multiple objects of the same class. They all contribute to the classification, which makes the object heat maps divergent (see the 7th column in Fig. \ref{image_heat}).

\begin{table}[t]
\caption{Average mEAS per class versus the correct localization precision (CorLoc \cite{deselaers2012weakly}) on PASCAL VOC 2007 \emph{trainval} split. The correlation coefficient of these two variables is $0.703$. } \label{meas_loc}
\vspace{0.1cm}
\footnotesize
\setlength\tabcolsep{8.2pt}
\begin{tabular}{|l|c|c||l|c|c|}
\hline
Class  &mEAS &CorLoc  &Class&mEAS &CorLoc    \\ \hline
bus   &0.306&0.699&car   &0.262&0.750 \\ \hline
tv    &0.254&0.582&aero  &0.220&0.685 \\ \hline
mbike &0.206&0.829&train &0.206&0.628 \\ \hline
horse &0.195&0.672&cow   &0.185&0.681 \\ \hline
boat  &0.177&0.343&sheep &0.176&0.719 \\ \hline
bike  &0.170&0.675&bird  &0.170&0.567 \\ \hline
sofa  &0.165&0.620&plant &0.163&0.437 \\ \hline
person&0.162&0.288&bottle&0.150&0.328 \\ \hline
cat   &0.143&0.457&dog   &0.135&0.406 \\ \hline
chair &0.093&0.171&table &0.052&0.305 \\ \hline
\end{tabular}
\vspace{-0.1cm}
\end{table}
%In practice, mEAS lowers image difficulty scores in this two situations and avoid picking inaccurate objects from these images for the following training.
%These images usually have a relatively clean background, and the targets are rigid objects, making it easy to localize the whole object as discriminative details. This usually happens in situations where the background is cluttered (\emph{chair} and \emph{dingtable}), or the targets are none rigid objects (\emph{cats} and \emph{person}), making it hard to pick out the whole objects as distinctive features. the difficulty of an image as how hard it is for a human to decide the presence or absence of a given object class in an image. Our difficulty measurement is model oriented, \emph{i.e.,} given image labels, we quantify the difficulty as the reliability of a model localizing the object of interest. As we can see, for a network, the successful detection results from the relatively clean background, and is not related with the target object size. This is different from
%Note that this is different from \cite{tudor2016hard}, the difficulty score is category specific, \emph{i.e.}, image may be difficult for localizing person, but easy to localize car. We consider the difficulty of an image as how hard it is for the network to decide of a given object in an image.

In addition, based on the mEAS, we are also able to analyze image difficulty at the class level. We compute mEAS at the class level by averaging the scores of images that contain the target object. In Table \ref{meas_loc}, we show the difficulty scores for all the 20 categories on PASCAL VOC 2007 \emph{trainval} split, along with the localization performance \cite{Bilen_2016_CVPR} in terms of CorLoc \cite{deselaers2012weakly}. We find that mEAS is highly related with the localization precision, with a correlation coefficient as high as $0.703$. In this dataset,  \textit{chair} and \textit{table} are the most difficult classes, containing cluttered scenes or partial occlusion. On the other hand, rigid objects such as \textit{bus} and \textit{car} are the easiest to localize, because these objects are usually large in images, or in relatively clean background.

\subsection{Progressive Detection Network}
Given the image difficulty scores and the mined seed positive instances, we are able to organize our network training in a progressive learning mode. The  detection network follows a fast-RCNN \cite{girshick2015fast} framework. Specifically, we split the training images $\mathcal{D}$ into $K$ folds $\mathcal{D}=\{\mathcal{D}_1,...,\mathcal{D}_K\}$, which are in an easy-to-difficult order. Instead of training and relocalization on the entire images all at once, we progressively recruit samples in terms of image difficulty. The training process starts with running a fast-RCNN on the first fold $\mathcal{D}_1$, which contains the easiest images, and obtains a trained model $M_{\mathcal{D}_1}$. $M_{\mathcal{D}_1}$ already has a good generalization ability since the trained object instances are highly reliable. Then we move on to the second fold $\mathcal{D}_2$, which contains relatively more difficult images. Instead of performing training and relocalization from scratch, we choose the trained model $M_{\mathcal{D}_1}$ to discover object instances in fold $\mathcal{D}_2$. It is likely to find more reliable instances on $\mathcal{D}_1 \bigcup \mathcal{D}_2$. As the training process proceeds, more images are added in, which improves the localization ability of the network steadily. When reaching later folds, the learned model has been powerful enough for localizing these difficult images.
%The system takes an image and the corresponding region proposals $\mathcal{R}$ (generated with edge boxes \cite{zitnick2014edge}) as inputs. The image is first passed through a series of convolutional (\emph{conv}) and max-pooling layers, resulting in feature maps in the last \emph{conv} block (noted as \emph{conv5} layer). Similar with \cite{girshick2015fast}, these feature maps are fed into a region of interest (RoI) pooling layer, which aims to map each region proposal to a fixed-length feature vector. These vectors are passed through another two fully connected (\emph{fc}) layers, and branched into two loss layers: one is a discrete probability distribution over $C+1$ categories (plus ``background''), and the other one is bounding box regression offsets. The improvements are two folds, first, we propose a self paced learning strategy based on image difficulty scores, and second, a weighted loss is introduced to model output loss, which consider the reliability of the mined instances.

%The images with high mEAS are first feeded into the network, hoping that they offer relatively reliable seed positive instances, and obtain a

\begin{algorithm}[t]
\caption{Zigzag Learning based Weakly Supervised Detection Network}\label{algorithm1}
\begin{algorithmic}
\REQUIRE  Training set $\mathcal{D} = \{x_i\}_{i=1}^N$ with image-level labels $Y=\{y_i\}_{i=1}^N$, iteration folds $K$, and masking ratio $\tau$; \\
\textbf{Estimating Image Difficulty:} Given an image $x$ with label $y \in \{1,-1\}^C$ and region proposals $\mathcal{R}$:\\
\quad i). Obtain region scores $x_{cr}\!\in\!{R^{C\times\mathcal{|R|}}}$ with WSDDN.\\
\quad ii). For each $y_c=1$, compute $\textit{mEAS}(X_c)$ with Eq. (\ref{meas}), \\
\quad and the object instance $x_{c}^o$ with Eq. (\ref{heatmap}).\\
\textbf{Progressive Learning:} Divide $\mathcal{D}$ into $K$ folds $\mathcal{D}=\{\mathcal{D}_1,...,\mathcal{D}_K\}$ according to mEAS.\\
\textbf{for} fold $k=1$ to $K$ \textbf{do}\\
\quad i). \textbf{Training} detection model $M_k$ with current selec-\\
\quad tion of object instances in $\bigcup_{i=1}^k \mathcal{D}_i$, \\
 \quad \quad a). given an image $x$, compute the last convolutional \\
  \quad \quad feature maps $\phi(x,f_{conv})$. \\
 \quad \quad b). for each mined object instance $x_c^o$, randomly se-\\
  \quad \quad lect regions $\{\Omega|\frac{S_{\Omega}}{S_{x_c^o}}=\tau\} $, and set $\phi(\Omega, f_{conv})=0$. \\
\quad  \quad c). continue forward and back propagation. \\
\quad ii). \textbf{Relocalize} object instances in folds $\bigcup_{i=1}^{k+1} \mathcal{D}_i$ using \\
 \quad current detection model $M_k$:\\
 \textbf{end for}
\ENSURE Detection models $\{M_k\}_{k=1}^K$.
\end{algorithmic}
\end{algorithm}

\textbf{Weighted loss.} Due to the high variation of image difficulty, the mined object instances used for training cannot be all reliable. It is suboptimal to treat all these instances equally important. Therefore, we penalize the output layers with a weighted loss, which considers the reliability of the mined instances. At each relocalization step, the network $M_k$ returns a detection score for each region, indicating its confidence of containing the target object. Formally, let $x^o_c$ be the relocalized object with instance label $y^o_c\!=\!1$, and $\phi^c(x^o_{c},M_k)$ be the detection score returned by $M_k$. The weighted loss w.r.t. region $x^o_c$ in the next retraining step is defined as
\begin{equation}\label{reg_loss}
\!L_{cls}(x^o_c,y^o_c,M_{k\!+\!1})\!=\!-\phi^c(x^o_{c},M_k)\log\phi^c(x^o_c, M_{k\!+\!1}).
\end{equation}
%where $\sigma$ is a softplus function controlling the decay factor of confidence with respect to detection score.

\subsection{Convolutional Feature Masking Regularization}
The above detector learning proceeds by alternating between model retraining and object relocalization, and is easy to get stuck in sub-optimums without proper initialization. Unfortunately, due to lack of object annotations, the initial seeds inevitably include inaccurate samples. As a result, the network tends to overfit those inaccurate instances during each iteration, leading to poor generalization. To solve this issue, we propose a regularization strategy to avoid the network from overfitting initial seeds in the proposed zigzag learning. Concretely, during network training, we randomly mask out those discriminative details at previous training, which enforces the network to focus on those less discriminative details, so that the current network can see a more holistic object.

The convolutional feature masking operation works as follows. Given an image $x$ and the mined object $x_c^o$ for each $y_c\!=\!1$, we randomly select region $\Omega \in x_c^o$ with $S_{\Omega}/S_{x_c^o}=\tau$, where $S_{\Omega}$ denotes the area of region $\Omega$. As $x_c^o$ obtains the highest responses during previous iteration, $\Omega$ is among the most discriminative regions. For each pixel $[u,v] \in \Omega$, we project it onto the last convolutional feature maps $\phi(x,f_{conv})$, such that the pixel $[u,v]$ in the image domain is closest to the receptive field of that feature map pixel $[u',v']$. This mapping is complicated due to the padding operations among convolutional and pooling layers. To simplify the implementation, following \cite{he2014spatial}, we pad $\left \lfloor p/2 \right \rfloor$ pixels for each layer with a filter size of $p$. This establishes a rough correspondence between a response centered at $[u', v']$, and receptive field in the image domain centered at $[Tu',Tv']$, where $T$ is the stride from the image to the target convolutional feature maps. The mapping of $[u, v]$ to the feature map $[u',v']$ is simply conducted as
\begin{equation}\label{mapping}
  u'=round((u\!-\!1)/T\!+\!1), v'=round((v\!-\!1)/T\!+\!1).
\end{equation}
In our experiments, $T=16$ for all models. During each iteration, we randomly mask out the regions by setting $\phi(\Omega, f_{conv})=0$, and continue forward and backward propagation as usual. For simplicity, we keep the aspect ratio of the masked region $\Omega$ the same as the mined object $x_c^o$. The whole process is summarized in Algorithm \ref{algorithm1}.
%with the mined object instance $x_o^c = [u_0, v_0, u_1, v_1]$ for category $c$, where $[u_0, v_0], \, [u_1, v_1]$ denote the upper left and bottom right coordinates of $x_o$, respectively. Based on , we choose the most discriminative parts within $x_o^c$

\section{Experiments}
We evaluate our proposed zigzag learning for weakly supervised object detection, providing extensive ablation studies and making comparison with state-of-the-arts.
\subsection{Experimental Setup}
\noindent \textbf{Datasets and evaluation metrics.} We evaluate our approach on PASCAL VOC 2007 \cite{everingham2010pascal} and 2012 \cite{everingham2015pascal} datasets. The VOC 2007 contains a total of 9,963 images spanning 20 object classes, of which 5,011 images are used for \emph{trainval} and the rest 4,952 images for \emph{test}. The VOC 2012 contains 11,540 images for \emph{trainval} and 10,991 images for \emph{test}. We choose the \emph{trainval} split for network training. For performance evaluation, two kinds of measurements are used: 1) CorLoc \cite{deselaers2012weakly} evaluated on the \emph{trainval} split; 2) the VOC protocol which measures the detection performance with average precision (AP) on the \emph{test} split.
%CorLoc measures the percentage of images with correct localization, where a window is considered to be correct if it has an IoU ratio of at least $50\%$ with one of the ground truth instances.

\noindent \textbf{Implementation details.}
We choose two CNN models to evaluate our approach: 1) CaffeNet \cite{jia2014caffe}, which we refer to as model \textbf{S} (meaning ``small''), and 2) VGG-VD \cite{Simonyan14c} (the $16$-layer model is used), which we call model \textbf{L} (meaning ``large''). In progressive learning, the training is run for $12$ epoches for each iteration, with learning rate $10^{-4}$ for the first $6$ epoches and  $10^{-5}$ for the last $6$ epoches. We choose edge boxes \cite{zitnick2014edge} to generate $|\mathcal{R}|\!\approx\!2000$ region proposals per image on average. All experiments use \emph{single-scale} ($s\!=\!600$) for training and test. We denote the length of its shortest side as the scale $s$ of an image. For data augmentation, we regard all proposals that have IoU $\geq 0.5$ with the mined objects as positive. The proposals that have IoU $\in [0.1,0.5)$ are treated as hard negative samples.The mean outputs of the $K$ models $\{M_k\}_{k=1}^K$ are chosen for test.

%The iteration fold $K$ and masking ratio $\tau$ are obtained via cross validation on PASCAL VOC 2007 \emph{trainval} split, and we follow the settings ($K=3$, $\tau=0.1$) in all experiments.
%We choose MatConvNet toolbox \cite{vedaldi15matconvnet} to conduct our experiments.
\begin{figure}[t]
  \centering
  % Requires \usepackage{graphicx}
  \includegraphics[width=0.45\textwidth]{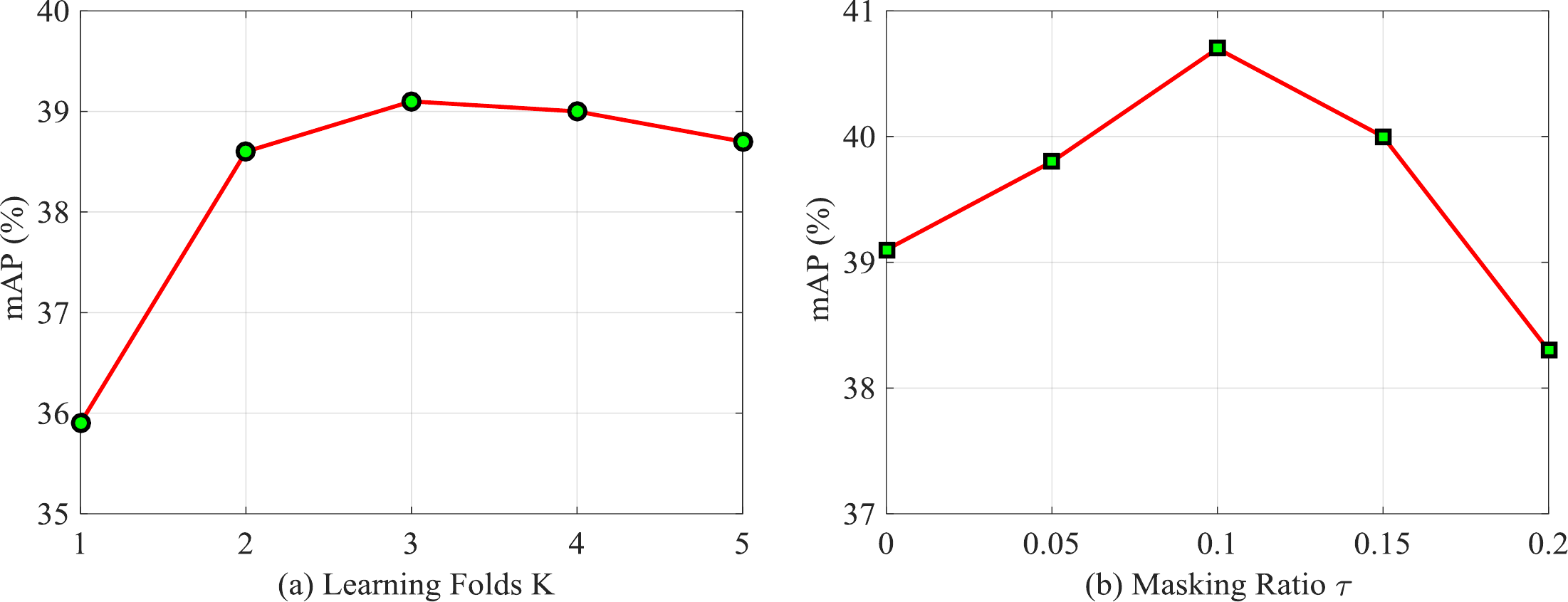} \\
   \vspace{0.1cm}
  \caption{Detection performance on PASCAL VOC 2007 \emph{test} split for different learning folds $K$ (left) and masking ratio $\tau$ (right).} \label{ablation}
  \vspace{-0.1cm}
\end{figure}
\subsection{Ablation Studies}
We first analyze the performance of our approach with different configurations. Then we evaluate the localization precision of different folds to validate the effectiveness of the mEAS. At last, we analyze the influences of two parameters: the progressive learning folds $K$ and the masking ratio $\tau$. Without loss of generality, all experiments here are conducted on PASCAL VOC 2007 with model \textbf{S}.

\begin{table}[t]
\caption{Detection performance comparison of  model \textbf{S} with various configurations on PASCAL VOC 2007 \emph{test} split.} \label{step_prob}
\setlength\tabcolsep{6.8pt}
\vspace{0.2cm}
\begin{tabular}{l|cccc}
& \multicolumn{4}{c}{Model \textbf{S}} \\ \hline
Region Scores?  & $\bm{\surd}$ & & & \\ \hline
mEAS ? & & $\bm{\surd}$&$\bm{\surd}$&$\bm{\surd}$ \\ \hline
Weighted Loss?& & &$\bm{\surd}$&$\bm{\surd}$ \\ \hline
Random Mask?  & & & & $\bm{\surd}$ \\ \hline
VOC 07 mAP& 34.1\% &37.7\% &39.1\% & 40.7\%  \\
\end{tabular}
\vspace{-0.2cm}
\end{table}

\begin{figure*}[!t]
  \centering
  % Requires \usepackage{graphicx}
  \includegraphics[width=0.99\textwidth, height=4.2cm] {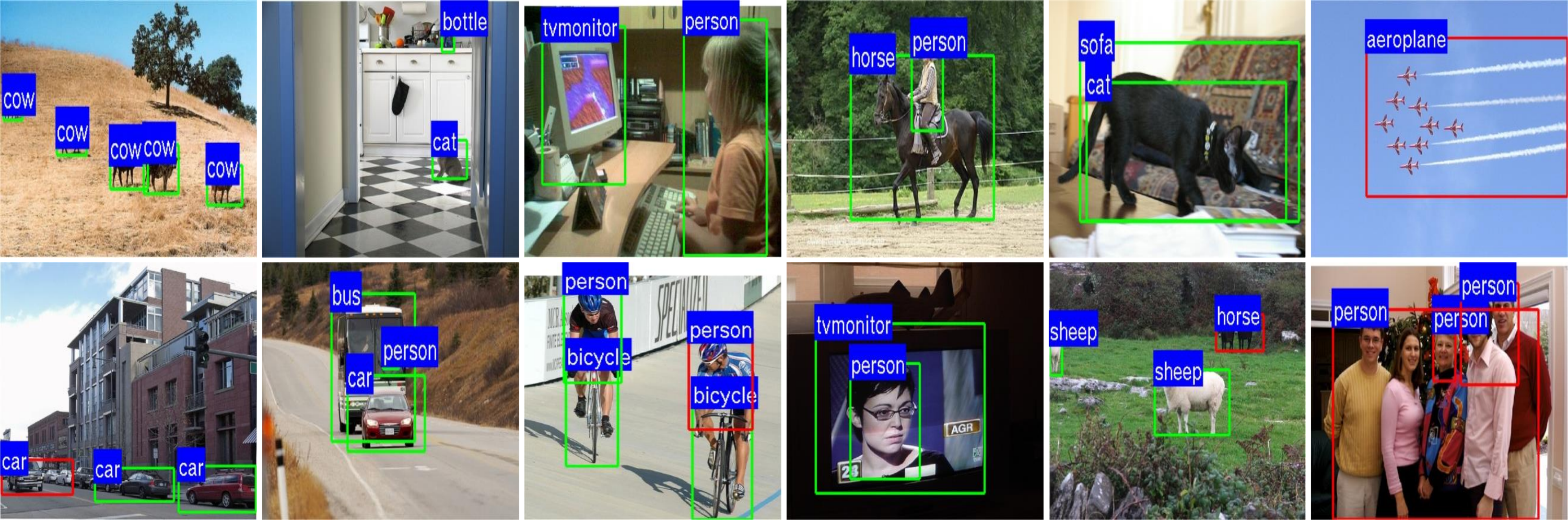} \\
  \vspace{0.1cm}
  \caption{Example detections on PASCAL VOC 2007 \emph{test} split ($47.6\%$ mAP). The successful detections (IoU $\geq 0.5$) are marked with green bounding boxes, and the failed ones are marked with red. We show all detections with scores $\geq 0.7$ and use nms to remove duplicate detections. The failed detections often come from localizing object parts or grouping multiple objects from the same class.}\label{detection}
  %For visualization, we set the detection threshold as $0.7$, and use NMS to remove duplicate detections.
  \vspace{0.1cm}
\end{figure*}

\begin{table*}[!t]
\caption{Localization precision ($\%$) on PASCAL VOC 2007 \emph{trainval} split at different fold iterations. The number of total folds is $K\!=\!3$.} \label{loc_results_07_fold}
\small
\setlength\tabcolsep{2.6pt}
\begin{tabular}{l|cccccccccccccccccccc|c}
\hline
Fold&aero&bike&bird&boat&bottle&bus&car&cat&chair&cow&
table&dog&horse&mbike&persn&plant&sheep&sofa&train&tv&mean \\ \hline
k=1 &87.3&90.0&81.8&56.7&69.1&85.5&88.9&62.5&27.0&
80.9&61.2&53.2&85.4&92.6&36.1&62.7&78.1
&81.6&79.3&85.9& 72.3 \\ \hline
k=2 &72.5&76.8&60.9&23.0&20.7&67.7&83.2&61.1&12.8&
78.7&48.5&51.8&74.8&88.9&27.4&35.4&64.5&54.6&63.4&67.4& 56.8 \\ \hline
k=3 &64.6&40.7&38.2&28.3&24.7&46.8&68.8&58.0&7.4&
55.3&26.9&58.2&58.3&77.1&30.2&27.7&51.5&44.7&32.2&45.9& 44.3 \\ \hline
\end{tabular}
\vspace{0.1cm}
\end{table*}

\begin{table*}[!t]
\caption{Localization precision ($\%$) on PASCAL VOC 2007 \emph{trainval} split in terms of CorLoc \cite{deselaers2012weakly} metric.} \label{loc_results_07}
\small
\setlength\tabcolsep{1.8pt}
\begin{tabular}{l|cccccccccccccccccccc|c}
\hline
method&aero&bike&bird&boat&bottle&bus&car&cat&chair&cow&
table&dog&horse&mbike&persn&plant&sheep&sofa&train&tv&mean \\ \hline
%Mimick \cite{li2016image}
%&73.1&45.0&43.4&27.7&6.8&53.3&58.3&45.0&6.2&48.0&14.3&47.3&69.4&66.8&24.3&
%12.8&51.5&25.5&65.2&16.8&40.0 \\ \hline
PLSA \cite{wang2014weakly}
&80.1&63.9&51.5&14.9&21.0&55.7&74.2&43.5&26.2&53.4&16.3&56.7&58.3&69.5
&14.1&38.3&58.8&47.2&49.1&60.9&48.5 \\ \hline
WSDDN \cite{Bilen_2016_CVPR} &65.1&58.8&58.5&33.1&39.8&68.3&60.2&59.6&34.8&64.5&30.5
&43.0&56.8&82.4&25.5&41.6&61.5&55.9&65.9&63.7&53.5 \\ \hline
PDA \cite{Li_2016_CVPR} &78.2&67.1&61.8&38.1&36.1&61.8&78.8&55.2&28.5
&68.8&18.5&49.2&64.1&73.5&21.4&47.4&64.6&22.3&60.9&52.3
&52.4 \\ \hline
DSD \cite{jie2017deep} &72.7&55.3&53.0&27.8&35.2&68.6&81.9&\textbf{60.7}&11.6&71.6&29.7
&\textbf{54.3}&64.3&88.2&22.2&53.7&72.2&52.6&68.9&75.5&56.1 \\ \hline
OICR \cite{Tang_2017_CVPR} &\textbf{81.7}&\textbf{80.4}&48.7&\textbf{49.5}&32.8&\textbf{81.7}&\textbf{85.4}
&40.1&\textbf{40.6}&\textbf{79.5}&35.7&33.7&60.5&\textbf{88.8}&21.8&\textbf{57.9}
&\textbf{76.3}&59.9&\textbf{75.3}&\textbf{81.4}&60.6 \\ \hline
ZLDN-\textbf{S}
&74.8&69.1&60.3&35.9&38.1&66.7&80.2&60.5&15.7&71.6&45.5
&54.4&72.8&86.1&\textbf{31.2}&42.0&64.6&60.3&58.6&66.4&57.8 \\ \hline
ZLDN-\textbf{L}
&74.0&77.8&\textbf{65.2}&37.0&\textbf{46.7}&75.8&83.7&58.8&17.5&73.1&\textbf{49.0}
&51.3&\textbf{76.7}&87.4&30.6&47.8&75.0&\textbf{62.5}&64.8&68.8&\textbf{61.2} \\ \hline
\end{tabular}
\vspace{0.1cm}
\end{table*}

\begin{table*}[!t]
\caption{Detection average precision ($\%$) on PASCAL VOC 2007 \emph{test} split.} \label{det07_results}
\small
\setlength\tabcolsep{1.6pt}
\begin{tabular}{l|cccccccccccccccccccc|c}
\hline
method&aero&bike&bird&boat&bottle&bus&car&cat&chair&cow&
table&dog&horse&mbike&persn&plant&sheep&sofa&train&tv&mAP \\ \hline
%MIL \cite{cinbis2014multi}
%&39.3&43.0&28.8&20.4&8.0&45.5&47.9&22.1&8.4&33.5&23.6&29.2&38.5&47.9
%&20.3&20.0&35.8&30.8&41.0&20.1&30.2 \\ \hline
pLSA \cite{wang2014weakly}
&48.8&41.0&23.6&12.1&11.1&42.7&40.9&35.5&11.1&36.6&18.4&35.3&34.8&51.3
&17.2&17.4&26.8&32.8&35.1&45.6&30.9 \\ \hline
WSDDN \textbf{S} \cite{Bilen_2016_CVPR} &42.9&56.0&32.0&17.6&10.2&61.8&50.2&29.0&3.8&36.2&18.5&31.1&45.8&54.5
&10.2&15.4&36.3&45.2&50.1&43.8&34.5 \\ \hline
WSDDN \textbf{L} \cite{Bilen_2016_CVPR} &39.4&50.1&31.5&16.3&12.6&64.5&42.8&42.6&10.1&35.7&24.9&38.2&34.4&55.6
&9.4&14.7&30.2&40.7&54.7&46.9&34.8 \\ \hline
PDA \cite{Li_2016_CVPR} &54.5&47.4&41.3&20.8&17.7&51.9&63.5&46.1&
21.8&57.1&22.1&34.4&50.5&61.8&16.2&\textbf{29.9}&40.7&15.9&55.3&40.2&39.5 \\ \hline
DSD \cite{jie2017deep} &52.2&47.1&35.0&\textbf{26.7}&15.4&61.3&66.0&54.3&3.0&53.6&24.7
&\textbf{43.6}&48.4&65.8&6.6&18.8&51.9&43.6&53.6&62.4&41.7 \\ \hline
OICR \cite{Tang_2017_CVPR} &\textbf{58.0}&62.4&31.1&19.4&13.0&\textbf{65.1}&62.2&28.4&\textbf{24.8}&44.7&30.6
&25.3&37.8&65.5&15.7&24.1&41.7&46.9&\textbf{64.3}&\textbf{62.6}&41.2 \\ \hline
ZLDN-\textbf{S} &51.9&57.5&40.9&15.8&17.6&53.3&61.2&54.0&2.0&44.2&42.9&34.5&58.3
&60.3&18.8&20.7&44.9&43.4&43.5&48.3&40.7 \\ \hline
ZLDN-\textbf{L} &55.4&\textbf{68.5}&\textbf{50.1}&16.8&\textbf{20.8}&62.7&\textbf{66.8}&\textbf{56.5}
&2.1&\textbf{57.8}&\textbf{47.5}&40.1&\textbf{69.7}&\textbf{68.2}&\textbf{21.6}
&27.2&\textbf{53.4}&\textbf{56.1}&52.5&58.2&\textbf{47.6} \\ \hline
\end{tabular}
\vspace{0.1cm}
\end{table*}

\begin{table*}[!t]
\caption{Localization precision ($\%$) on PASCAL VOC 2012 \emph{trainval} split in terms of CorLoc \cite{deselaers2012weakly} metric.} \label{loc_results_12}
\small
\setlength\tabcolsep{2.0pt}
\begin{tabular}{l|cccccccccccccccccccc|c}
\hline
method&aero&bike&bird&boat&bottle&bus&car&cat&chair&cow&
table&dog&horse&mbike&persn&plant&sheep&sofa&train&tv&mean \\ \hline
DSD \cite{jie2017deep} &82.4&68.1&54.5&38.9&35.9&\textbf{84.7}&73.1&\textbf{64.8}&17.1&78.3&22.5
&\textbf{57.0}&\textbf{70.8}&86.6&18.7&49.7&80.7&45.3&70.1&77.3&58.8 \\ \hline
OICR \cite{Tang_2017_CVPR} &\textbf{86.2}&\textbf{84.2}&\textbf{68.7}&\textbf{55.4}&46.5&82.8&74.9
&32.2&\textbf{46.7}&\textbf{82.8}&\textbf{42.9}&41.0&68.1&\textbf{89.6}&9.2&53.9
&\textbf{81.0}&52.9&59.5&\textbf{83.2}&\textbf{62.1} \\ \hline
ZLDN-\textbf{L} &80.3&76.5&64.2&40.9&\textbf{46.7}&78.0&\textbf{84.3}&57.6&21.1&
69.5&28.0&46.8&70.7&89.4&\textbf{41.9}&\textbf{54.7}&76.3
&\textbf{61.1}&\textbf{76.3}&65.2& 61.5 \\ \hline
\end{tabular}
\vspace{0.1cm}
\end{table*}

\begin{table*}[!t]
\caption{Detection average precision ($\%$) on PASCAL VOC 2012 \emph{test} split.} \label{det12_results}
\small
\setlength\tabcolsep{2.0pt}
\begin{tabular}{l|cccccccccccccccccccc|c}
\hline
method&aero&bike&bird&boat&bottle&bus&car&cat&chair&cow&
table&dog&horse&mbike&persn&plant&sheep&sofa&train&tv&mAP \\ \hline
PDA \cite{Li_2016_CVPR}
 &62.9&55.5&\textbf{43.7}&14.9&13.6&57.7&52.4&50.9&13.3&45.4&4.0&30.2&55.6&67.0&3.8&23.1
 &39.4&5.5&50.7&29.3&35.9 \\ \hline
DSD \cite{jie2017deep} &60.8&54.2&34.1&14.9&13.1&54.3&53.4&\textbf{58.6}&3.7&\textbf{53.1}&8.3
&\textbf{43.4}&49.8&\textbf{69.2}&4.1&17.5&43.8&25.6&\textbf{55.0}&50.1&38.3 \\ \hline
OICR \cite{Tang_2017_CVPR} &\textbf{67.7}&61.2&41.5&\textbf{25.6}&\textbf{22.2}&54.6&49.7&25.4&\textbf{19.9}&47.0&18.1
&26.0&38.9&67.7&2.0&22.6&41.1&34.3&37.9&\textbf{55.3}&37.9 \\ \hline
ZLDN-\textbf{L} &54.3&\textbf{63.7}&43.1&16.9&21.5&\textbf{57.8}&\textbf{60.4}&50.9&1.2&
51.5&\textbf{44.4}&36.6&\textbf{63.6}&59.3&\textbf{12.8}&\textbf{25.6}&\textbf{47.8}
&\textbf{47.2}&48.9&50.6& \textbf{42.9} \\ \hline
\end{tabular}
\end{table*}

\noindent $\bullet$ \textbf{Component analysis.} To reveal the contribution of each module, we test the detection performance with different configurations. These variants include: 1) using region scores (Eq.~(\ref{region_score})) as image difficulty metric; 2) using the proposed mEAS for image difficulty measurement; 3) introducing weighted loss during model retraining; and 4) adding masking regularization. The results are shown in Table \ref{step_prob}. From the table we observe the following three aspects.

1) The mEAS is more effective than region scores from
Eq.~(\ref{region_score}), with a gain up to about $3.2\%$ ($34.1\% \rightarrow 37.7\%$). The main reason is as follows. For deformable objects like \emph{bird} and \emph{cat}, the highest region scores may focus on object parts, thus the progressive learning chooses inaccurate object instances during initial training. In contrast, mEAS lowers those scores only concentrating on part of the objects by introducing convergent measurement, and avoids choosing these parts for initial detector training.

2) Introducing weighted loss brings about $1.4\%$ gain. This demonstrates that considering the confidence of the mined object instances helps boost the performance.

3) The proposed masking strategy further boosts the performance to an accuracy of $40.7\%$, which is $1.6\%$ better than the baseline. This demonstrates that the masking strategy can effectively prevent the model from ovetfitting and enhance its generalization ability.

\noindent $\bullet$ \textbf{CorLoc versus fold iteration.} In order to validate the effectiveness of mEAS, we test the localization performance during each iteration in terms of CorLoc. Table~\ref{loc_results_07_fold} shows the localization results on VOC 2007 \emph{trainval} split when learning folds $K\!=\!3$. During the first iteration ($k=1$) for the easiest images, our method achieves an accuracy of $72.3\%$. When moving on to more difficult images ($k=2$), the performance is decreased to $56.8\%$.  It only achieves $44.3\%$ for the most difficult image fold, even though we have a more powerful model when $k=3$. The results demonstrate that mEAS is an effective criterion to measure the difficulty of an image w.r.t. localizing the corresponding object.
%In Fig. ~(\ref{image_heat}) we illustrate some examples to uncover the relationship between mEAS and the localization results. In this subsection, we quantify their relatedness in terms of CorLoc \cite{deselaers2012weakly}.

\noindent $\bullet$ \textbf{Learning folds $K$.} Fig. \ref{ablation}(a) shows the detection results w.r.t. different learning folds, where $K\!=\!1$ means that the training process chooses entire images all at once, without using progressive learning. We find that the progressive learning strategy significantly improves the detection performance. The result is $39.1\%$ for $K\!=\!3$, \ie about $3.2\%$ gain over the baseline ($35.9\%$). The performance tends to be saturated as $K$ increases and even slightly drops, mainly because too few images in initial stages degrade the model's detection power.

\noindent $\bullet$ \textbf{Masking ratio $\tau$.} The masking ratio $\tau$ denotes the percentage of area $\Omega$ versus that of the mined object $x_c^o$. Fig. \ref{ablation}(b) shows the results as we mask out different ratios of the mined objects. With masking ratio $\tau\!=\!0.1$, the test performance reaches $40.7\%$, which surpasses the baseline without using masking by $1.6\%$. The improvement demonstrates that the proposed masking strategy is able to enhance the generalization ability of the trained model. As the masking ratio increases, the performance gradually drops, mainly because masking too many regions prevents the model from seeing true positive samples.

\begin{figure}[!t]
  \centering
  % Requires \usepackage{graphicx}
  \includegraphics[width=0.46\textwidth]{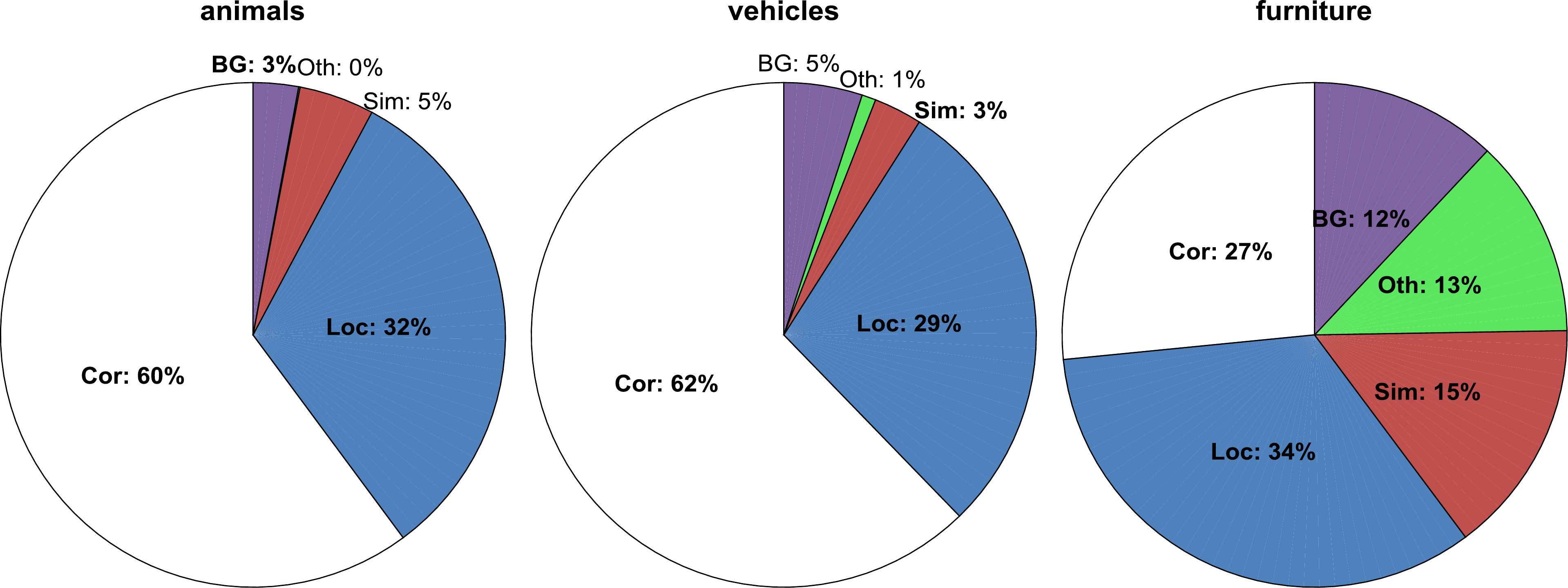} \\
  \vspace{0.1cm}
  \caption{Detection error analysis \cite{hoiem2012diagnosing} of our proposed model on animals, vehicles, and furniture from VOC 2007 \emph{test} split. The detections are categorized as correct (Cor), false positive due to poor localization (Loc), confusion with similar categories (Sim), with others (Oth), and with background (BG).} \label{error_analysis}
  %The detections are categorized as correct (Cor), false positive due to poor localization (Loc), confusion with similar categories (Sim), with others (Oth), and with background (BG).
  \vspace{-0.1cm}
\end{figure}

\subsection{Comparisons with state-of-the-arts} We then compare our results with state-of-the-arts for weakly supervised detection. Our method is denoted as \textbf{ZLDN}, standing for Zigzag Learning Detection Network. Unless specified, all other results are based on model \textbf{L}.

\noindent $\bullet$ \textbf{CorLoc evaluation.} Table \ref{loc_results_07} shows the localization results on PASCAL VOC 2007 \emph{trainval} split in terms of CorLoc \cite{deselaers2012weakly}. Comparing with WSDDN \cite{bency2016weakly} ($53.5\%$), our method brings $7.7\%$ improvement, this mainly results from the zigzag learning. Our method achieves slightly better localization performance ($61.2\%$) compared with previous best-performing method \cite{Tang_2017_CVPR} ($60.6\%$). Similar results can be found in Table \ref{loc_results_12} which shows the localization performance on VOC 2012. Our method obtains an accuracy of $61.5\%$, which is comparable with the best performing method \cite{Tang_2017_CVPR} ($62.1\%$). Note that the result of \cite{Tang_2017_CVPR} is based on multiple scales, while our result is simply from the last learning iteration, which is in single scale.
%Another finding is that using deeper network brings about negligible improvement, one possible explanation is that for the deeper network, the recognition performance is improved by focusing on object parts, not the whole object.

\noindent $\bullet$ \textbf{AP evaluation.} Table \ref{det07_results} and Table \ref{det12_results} show the detection performance in average precision (AP) on PASCAL VOC 2007 and 2012 \emph{test} split, respectively. Just using model \textbf{S}, our method achieves an accuracy of $40.7\%$, \ie about $6.2\%$ improvement over the best-performing method WSDDN \cite{Bilen_2016_CVPR} ($34.5\%$) using the same model on VOC 2007. When switching to model \textbf{L}, the detection accuracy increases to $47.6\%$ on VOC 2007, which is about $6\%$ better than the best-performing result \cite{jie2017deep} ($41.7\%$). On PASCAL VOC 2012, the detection accuracy is $42.9\%$, which is $4.6\%$  better than previous state-of-the-art result \cite{jie2017deep} ($38.3\%$).

\noindent $\bullet$ \textbf{Error analysis and visualization.}  To show the performance of our model more detailedly, we use the analysis tool from \cite{hoiem2012diagnosing} to diagnose the detector error. Fig. \ref{error_analysis} shows the error analysis on PASCAL VOC 2007 \emph{test} split with model $\textbf{L}$ (mAP $47.6\%$). The classes are categorized into three categories, \emph{animals}, \emph{vehicles}, and \emph{furniture}. Our method achieves promising results on categories \emph{animals} and \emph{vehicles}, with an average precision above $60\%$, but it does not work well on detecting \emph{furniture}. This is mainly because \emph{furniture} like \emph{chair} and \emph{table} are usually in cluttered scenes, thus very hard to pick out for model training. On the other hand, the majority of error comes from inaccurate localization, which is around $30\%$ for all categories. We show some detection results in Fig. \ref{detection}. The correct detections are marked with green bounding boxes, while the failed ones are marked with red. It can be seen that the incorrect detections often come from detecting object parts, or grouping multiple objects from the same class.

Although our proposed method achieves better performance than previous works, it performs not very well on some categories, like \emph{chair} and \emph{person}. The reason is that the detection performance mainly dependents on the object instances obtained from the classification model, which is limited in correctly localizing these objects. Actually, localizing objects such as \emph{chair} and \emph{person} in cluttered backgrounds is the main challenge in weakly supervised detection, which remains  a further research direction.

\section{Conclusion}
This paper proposed a zigzag learning strategy for weakly supervised object detection. To develop such effective learning, we propose a new and effective criterion named mean Energy Accumulated Scores (mEAS) to automatically measure the difficulty of an image, and progressively recruit samples via mEAS for model training. Moreover, a masking strategy is incorporated into network training by randomly erasing the high responses over the last convolutional feature maps, which highlights the less discriminative parts and improves the network's generalization ability. Experiments conducted on PASCAL VOC benchmarks demonstrated the effectiveness of the proposed approach.

\noindent \small{\textbf{Acknowledgements}. The work was supported in part to Jiashi Feng by NUS startup R-263-000-C08-133, MOE Tier-I R-263-000-C21-112, NUS IDS R-263-000-C67-646 and ECRA R-263-000-C87-133, in part to Dr. Hongkai Xiong by NSFC under Grant 61425011, Grant 61720106001, Grant 61529101, and in part to Dr. Qi Tian by ARO grant W911NF-15-1-0290 and Faculty Research Gift Awards by NEC Laboratories of America and Blippar}.

%by the Program of Shanghai Academic Research Leader under Grant 17XD1401900,
%----------------------------------------------------------------------------------------------------------------------------------------------------
{\small
\bibliographystyle{ieeetr}
\bibliography{egbib}
}

\end{document}